\documentclass{article}

\usepackage{microtype}
\usepackage{graphicx}
\usepackage{subfigure}
\usepackage{booktabs} % for professional tables

\usepackage{hyperref}

\usepackage{hyperref}
\usepackage{url}
\usepackage{color}
\usepackage{multirow}
\usepackage{amssymb}
\usepackage{amsmath}
\usepackage{mathtools}

\usepackage[accepted]{icml2019arxiv}

\icmltitlerunning{Pseudo-Encoded Stochastic Variational Inference}

\begin{document}

\twocolumn[
\icmltitle{Pseudo-Encoded Stochastic Variational Inference}




\begin{icmlauthorlist}
\icmlauthor{Amir Zadeh}{lti}
\icmlauthor{Simon Hessner}{lti}
\icmlauthor{Yao-Chong Lim}{scs}
\icmlauthor{Louis-Philippe Morency}{lti}
\end{icmlauthorlist}

\icmlaffiliation{lti}{LTI, SCS, Carnegie Mellon University}
\icmlaffiliation{scs}{SCS, Carnegie Mellon University}
\icmlcorrespondingauthor{Amir Zadeh}{abagherz@cs.cmu.edu}

\icmlkeywords{Machine Learning, ICML}

\vskip 0.3in
]



\printAffiliationsAndNotice{} % otherwise use the standard text.

\begin{abstract}
Posterior inference in directed graphical models is commonly done using a probabilistic encoder (a.k.a inference model) conditioned on the input. Often this inference model is trained jointly with the probabilistic decoder (a.k.a generator model). If probabilistic encoder encounters complexities during training (e.g. suboptimal complxity or parameterization), then learning reaches a suboptimal objective; a phenomena commonly called inference suboptimality~\cite{cremer2018inference}. In Variational Inference (VI)\cite{jordan1999introduction}, optimizing the ELBo using Stochastic Variational Inference (SVI)~\cite{rezende2014stochastic} can eliminate the inference suboptimality (as demonstrated in this paper), however, this solution comes at a substantial computational cost when inference needs to be done on new data points. Essentially, a long sequential chain of gradient updates is required to fully optimize approximate posteriors. In this paper, we present an approach called Pseudo-Encoded Stochastic Variational Inference (PE-SVI), to reduce the inference complexity of SVI during test time. Our approach relies on finding a suitable initial start point for gradient operations, which naturally reduces the required gradient steps. Furthermore, this initialization allows for adopting larger step sizes (compared to random initialization used in SVI), which further reduces the inference time complexity. PE-SVI reaches the same ELBo objective as SVI using less than one percent of required steps, on average. 
\end{abstract}

\section{Introduction}

Training directed graphical models using Variational Inference (VI) has a long history in machine learning research~\citep{jordan1999introduction} . Commonly, inference is done using probabilistic inference models~\citep{dayan1995helmholtz} such as a probabilistic encoder in VAE~\citep{kingma2013auto}. Using a parameteric model to perform inference allows for fast inference given new datapoints. However, if inference network encounters difficulties, then  maximization of ELBo is done suboptimally~\citep{cremer2018inference}. Previous works have attempted to mitigate the inference suboptimality using fine-tuning~\citep{hjelm2016iterative}, ladder-based models~\citep{sonderby2016ladder} and Hessian-based models~\citep{kim2018semi}. While these attempts have been very successful in dealing with numerical instabilities, inference suboptimality due to limited inference model capacity is intertwined with the nature of inference models CITE. Alternatively, to avoid this inference suboptimality altogether, as shown in this paper, one can rely on Stochastic Variational Inference (SVI) using free-form posterior parameterization and mean-field approximation~\cite{rezende2014stochastic}. However, using SVI, inference for new datapoint requires a long gradient (or somewhat faster alternative meta-gradient approaches) update chain, which makes the inference suffer heavily during test time. Essentially, parameters of approxiamte posteriors are initialized randomly and updated iteratively until ELBo maximization objective is reached. 

In this paper, we assume the following separation about the inference process of SVI: 1) a suboptimal initial inference that a reasonably parameterized inference model can reach, 2) subsequent gradient-based updates to reach full ELBo maximization. Using the above assumption, we reach at a simple-yet-elegant framework called Pseudo-Encoded Stochastic Variational Inference (PE-SVI): a framework for test-time inference speed-up of SVI. The learning process is separated in three parts: (a) \textit{Early Decoder Training}: which trains a decoder using SVI to maximize the lower-bound of likelihood using tractable easy-to-sample approximate posteriors. (b) \textit{Deferred Encoder Training}: After the decoder and approximate posterior parameters are fully learned over the train set, a \textit{pseudo-encoder} is trained in a supervised fashion between input data points and their respective approximate posterior parameters. \textit{Pace Adjustment}: After initial approximate posterior parameter estimation using the trained encoder, the step size can be increased and tuned for fast convergence. Such large step sizes are often detrimental to SVI if approximate posterior parameters are initialized randomly. 

The following summarizes, contributions and findings of this paper:

\begin{itemize}
    \item We present a speed-up framework for test-time Stochastic Variational Inference (SVI), called Pseudo-Encoded Stochastic Variational Inference (PE-SVI). PE-SVI is easy to implement and does not require complex or costly calculations during train time (e.g. Hessian calculations \cite{kim2018semi}).
    \item PE-SVI is able to reach the same ELBo as SVI, with a fraction of the required steps. In our experiments over publicly available datasets, PE-SVI reaches similar performance as SVI in an average of $15.2$ gradient updates, while SVI takes substantially larger number of steps with an average of  $1826.1$. 
    \item To our surprise, ELBo loss achieved using PE-SVI's pseudo-encoder without any gradient steps in majority of times is better than end-to-end training of both encoder and decoder for VI (i.e. VAE). In simple terms, our experiments controversially hint that it is better to train the decoder first and subsequently the encoder, as opposed to training both end-to-end. This is further discussed in Section \ref{sec:results}. 
\end{itemize}
\section{Background and Related Works}
In this section we first start with the background required for VI and SVI. We subsequently discuss the comparison between our approach and previous methods for improving SVI inference complexity.

\subsection{Variational Inference}
Let samples drawn as $(z,x) \sim p(z) p(x|z)$ form a dataset $S=\{(z_i,x_i)\}_{i=1}^{|S|}$. $x_i,z_i$ are regarded as observed and latent variables. Unfortunately $z_i$, being the latent space generating the data $x_i$, is not observable. Therefore, MLE on the joint distribution is not possible. Considering a parametric distribution with parameters $\theta$, the marginal likelihood can be written as:
\begin{align}\label{eq:big}
\begin{split}
    \mathcal{L}^{(i)}(\theta) &=\log \int  p_\theta(z, x_i)\ dz = \\
    &-\int q_\phi(z|x_i)\ \log \ \frac{p_\theta(z|x_i)}{q_\phi(z|x_i)} \ dz  \quad \\ &+ \int q_\phi(z|x_i)\ \log \ \frac{p_\theta(z|x_i) p_\theta(x_i)}{q_\phi(z|x_i)} \ dz
\end{split}
\end{align}
Calculating the MLE using the first line of Equation \ref{eq:big} is still not tractable due to the latents being unobserved. Using Variational Inference (VI), a tractable and easy-to-sample approximate posterior distribution $q_\phi(\cdot)$ can be utilized as shown in the equation above. The second line of Equation \ref{eq:big} denotes two distinct terms with the condition that $q_\phi(z|x)>0 \iff p_\theta(z|x)>0$. The first term denotes the KL divergence between the real and approximate posterior distributions. Minimizing this KL term between parametric posterior $q_\phi(\cdot)$ and true posterior $p_\theta(\cdot)$ would allow for sampling from $q(\cdot)$ as proxy of $p_\theta(\cdot)$, however the KL cannot be efficiently calculated due to true posterior $p_\theta(\cdot)$ not being easy to sample from. The second term is the Evidence Lower Bound (ELBo) of the likelihood which is equal to the following:
\begin{equation}\label{eq:elbo}
    \textrm{ELBo} 
    =\mathbb{E}_{q_\phi(z|x_i)} [\log\ p_\theta(x|z)] - \textrm{KL} (q_\phi(z|x_i) || p_\theta(z))
\end{equation}

The first term in the RHS of Equation \ref{eq:elbo} is the expected reconstruction of the observed data using parametric probabilistic model $p_\theta$, under approximate density $q_\phi(\cdot)$. The seconds term encourages good prior density estimation for $q_\phi(\cdot)$, with the prior $p_\theta(\cdot)$ often being a desired distribution in practice. 

A notable neural model that follows the above variational framework is Variational Auto-Encoder (VAE \cite{kingma2013auto}). VAE uses an encoder to parameterize the distribution $q_\phi(\cdot)$. During learning, the AEVB algorithm is used for training an encoder (or inference network) and decoder jointly together using a reparameterization trick. An alternative framework is Stochastic Variational Inference (SVI \cite{hoffman2013stochastic}), where the approximate posterior parameterization is done using well-known distributions as opposed to a neural model. SVI has certain appealing applications, for example SVI framework is used in Variational Auto-Decoder (VAD) for learning generative models from data with severe missingness~\cite{zadeh2019variational}.

\subsection{Amortization Gap}

In a generative modeling framework, often the decoder is considered the main component of the model. This is conventionally the component that receives samples drawn from a latent posterior distribution, and generates new data points. Using SVI \cite{hoffman2013stochastic} with a mean-field assumption, one can train such a model without the need for an encoder (i.e. inference) network. However, if inference is ever required during test time, such models suffer heavily due to relying on test-time gradient (or meta-gradient) descent (which is a non-parallelizable sequential operation). Using an encoder allows the process of inference to become more efficient; during test time, one can simply feed the datapoint into an encoder to get the parameters of the posterior distribution. This process is far less computationally exhaustive than gradient-based inference (since operations inside a network are usually parallelized). 

However, if the inference network cannot be trained efficiently - e.g. has limited capacity, or undergoes difficulties during training, or simply the nature of data is too hard for dimensionality reduction using a neural structure - then the process of learning a generative model may be suboptimal. This is called an Amortization Gap~\citep{cremer2018inference}, which can be somewhat mitigated using methods that require second-order gradient of the model's optimizer~\citep{kim2018semi}. Amortization Gap is not a  theoretical weakness of inference networks (note the universal approximation theory of neural networks), but rather an empirical phenomena best describable by finite-neuron neural networks. A free-form mean-field approximate posterior inference technique such as SVI can mimic an encoder with very large capacity (due to mean-field assumption and full independence of latent parameters), and does not suffer the same gap (as shown in this paper). However, as mentioned, this comes at the cost of inference time complexity.

\section{Pseudo-Encoded Stochastic Variational Inference}
In this section, we outline the training process of the Pseudo-Encoded Stochastic Variational Inference (PE-SVI) framework. Training in PE-SVI is split into 3 parts: 1) Early Decoder Training, 2) Deferred Encoder Training, and 3) Pace Adjustment. 

\subsection{Early Decoder Training}

At the first step within PE-SVI framework, a decoder is trained (without an attached encoder). Essentially, we use Stochastic Variational Inference (SVI) with mean-field assumption on latent dimensions. Assume samples $z \sim q_\phi(z|x_i)$ are drawn from a given known family of distributions (e.g. Gaussian). To generate data similar to $x_i$, these samples are then used as input to a decoder $\mathcal{D}_{\theta}(z_i)$. The reconstructed samples of this probabilistic decoder should show high resemblance such that ELBo (Equation \ref{eq:elbo}) is maximized w.r.t $\theta$, and $\phi$. 

$p_{\theta}(x|z)$ in turn can be defined as: \vspace{2pt}
\begin{equation}
\label{eq:invdecoderp}
    p_{\theta}(x|z)=\mathcal{N}\left(\mathcal{D}_\theta(z); x_i, \Lambda_i \right)
\end{equation}

The high likelihood is therefore attributed to the low squared distance (as measure) between the output reconstruction of $\mathcal{D}_{\theta}(\cdot)$ and the ground-truth $x_i$. $\Lambda_i$ is the covariance matrix of the above likelihood. The approximate posterior is not parameterized by an encoder, but rather by well-known distributions such as a  Gaussian (in this paper): \vspace{3pt}
\begin{equation}\label{eq:qnormal}
q_\phi(z|x_i)\coloneqq \mathcal{N}(z; \mu_i,\Sigma_i)
\end{equation}

At the beginning of training, parameters of the approximate posterior $q_\phi(\cdot)$ are initialized randomly (e.g. uniform), same as parameters $\theta$ of the probabilistic decoder $p_\theta(\cdot)$.  Within each batch of the training data, the gradient of lower-bound is calculated and the parameters of $q_\phi(\cdot)$ and $p_\theta(\cdot)$ are updated. Since there is no encoder attached to the network, backpropagation is only done to the decoders parameters ($\theta$). In the meantime, backpropagation also happens for parameters of the approximate posterior ($\phi$). Updates on the parameters of approximate posterior $q_\phi(z|x_i)$ are only done once in an epoch, when backpropagating the ELBo for $x_i$. Training is done until convergence w.r.t both $\theta$ and $\phi$. The output of the Early Decoder Training phase is the trained approximate posteriors $q_{\phi^*}(z|x_i)$ as well as the trained decoder $\mathcal{D}_{\theta^*}(\cdot)$. 

\subsection{Deferred Encoder Training}\label{sec:latenc}

After training is done, we use a neural model, also referred to as pseudo-encoder in this paper, $\mathcal{E}_{\gamma}(x)$ to perform a similar role as an encoder. Unlike conventional encoder-decoder architectures (in which encoder is trained end-to-end alongside the decoder) in PE-SVI, the pseudo-encoder is trained only after decoder is fully learned. The learned approximate posteriors $q_{\phi^*}(z|x_i)$ of the Early Decoder Training phase are passed to Deferred Encoder Training phase, essentially to be approximated. The objective (and a supervised one at that) is to learn a mapping between $x_i$ and $\phi^*=\{\mu_i^*,\Sigma_i^*\}$. $\mathcal{E}_{\gamma}(x)$ is therefore trained in a supervised manner for this purpose, to output $\phi^*$ given $x_i$. This training can be done like any other supervision, using gradient descent approaches. After training is done $\mathcal{E}_{\gamma^*}(x)$ is used to provide a good estimate of the parameters of the true approximate posterior. For a datapoint $x_i$, we denote the estimates of the approximate posterior generated by $\mathcal{E}_{\gamma^*}(x)$ as $\phi^\mathcal{E}=\{\mu^\mathcal{E}_i, \Sigma^\mathcal{E}_i\}$.

\subsection{Pace Adjustment}\label{sec:adjustment}

For $i$th input $x_i$, the parameters of the approximate posterior $q_{\phi^\mathcal{E}}(z|x_i)$ are first obtained using the pseudo-encoder $\mathcal{E}_{\gamma^*}(x_i)$. Subsequently, these parameters can be refined using SVI to achieve the final posteriors $q_{\phi^*}(z|x_i)$. This by itself reduces the number of SVI steps required to maximize the ELBo to a significant amount (naturally due to approximation of the $\phi^*=\{\mu_i^*,\Sigma_i^*\}$ using $\phi^\mathcal{E}=\{\mu^\mathcal{E}_i, \Sigma^\mathcal{E}_i\}$, also shown in experiments in this paper). However, during Pace Adjustment phase, one can switch to SVI step\footnote{In this paper, we use Adam~\cite{kingma2014adam} as the optimizer for approximate posterior parameters.} sizes that are most suited for convergence, given the initial estimates of approximate posterior parameters $\phi^\mathcal{E}$. Therefore, higher learning rates, which are often detrimental if approximate posterior is initialized randomly, can be used to maximize ELBo w.r.t $\phi$ (initialized with $\phi^\mathcal{E}$). Thus, a further reduction in number of steps can be made by simply taking larger steps. One can simply treat the adjusted learning rate as a hyperparameter, and pick the one with fastest and most accurate convergence to $\phi^*$. Any hyperpatameter optimization method (e.g. Bayesian hyperparameter optimization approaches~\citep{snoek2012practical}) may be used for more accurate localization of a suitable pace. For the sake of this paper, we simply suffice to treating the adjusted learning rate as a hyperparameter found using random (yet sensible) grid search.

\section{Experiments}
In this section, we discuss the details of the experiments for this paper. We first start by discussing the studied datasets, followed by methodology and hyperparameter space search. 

\subsection{Datasets}
We use the following set of datasets in our experiments:

\textit{Synthetic Data:} As the first dataset in our experiment, we study a case of synthetic data where we control the distributional properties of the data. In the generation process, we first acquire a set of independent dimensions randomly sampled from 5 univariate distributions with uniform random parameters: \{\texttt{Normal, Uniform, Beta, Logistic, Gumbel}\}. Often in realistic scenarios there are inter-dependencies among the dimensions. Hence we proceed to generate interdependent dimensions by picking random subsets of the independent components and combining them using random operations such as weighted multiplication, affine addition, and activation. Using this method, we generate a dataset containing $N=50,000$ datapoints with ground-truth dimension $d=300$. 

\textit{CMU-MOSI Dataset:}  CMU Multimodal Opinion Sentiment Intensity (CMU-MOSI) is a dataset of multimodal language specifically focused on multimodal sentiment analysis~\citep{zadeh2016multimodal}. It is among the most well-studied multimodal language datasets in NLP community. Multimodal sentiment analysis extends conventional language-based sentiment analysis to a multimodal setup where both verbal and non-verbal signals contribute to the expression of sentiment. CMU-MOSI contains 2199 video segments taken from 93 Youtube movie review videos. The train, validation and test folds of the CMU-MOSI contain 1248, 229 and 686 segments respectively~\citep{chen2017multimodal}. We use expected multimodal context for each sentence, similar to unordered compositional approaches in NLP~\citep{iyyer2015deep}.

\textit{300-W:} \cite{Sagonas2013,Sagonas2013a} is a meta-dataset of four different facial landmark datasets: Annotated Faces in the Wild (AFW)~\citep{Zhu2012}, iBUG \citep{Tzimiropoulos2013}, and LFPW + Helen~\citep{Belhumeur2011,Le2012} datasets. We used the full iBUG dataset and the test partitions of LFPW and HELEN. This led to 135, 224, and 330 images for testing respectively. They all contain uncontrolled images of faces in the wild: in indoor-outdoor environments, under varying illuminations, in presence of occlusions, under different poses, and from different quality cameras. For the purpose of statistical shape modeling, only the landmarks are used. 

\textit{SST:} The Stanford Sentiment Treebank (SST) is a dataset of movie review excerpts from Rotten Tomatoes website~\citep{socher2013recursive}. The dataset is annotated for both root and intermediate nodes of parsed sentences. We only use the root nodes in our experiments. Similar to CMU-MOSI, we use an unordered compositional approach for the input sentence embeddings.
\begin{table}[t!]
\begin{center}
\small
%\fontsize{7}{10}\selectfont
\setlength{\tabcolsep}{6.2pt}
\renewcommand{\arraystretch}{1.4}
\begin{tabular}{l|c|c|c|c}
Model\ \textbackslash \ $|z|$ & 16 & 32 & 64 & 128 \\
\hline 
\multicolumn{5}{|c|}{CMU-MOSI} \\
\hline 
VAE & 0.7176 & 0.5871  & 0.4681 & 0.2623  \\
SVI & 0.0470 & 0.0010  & 0.0006 & 0.0003 \\
PE-SVI-0 & 0.0516 & 0.0064  & 0.0090 & 0.0063  \\
PE-SVI-25 &  0.0482 & 0.0010 & 0.0007 & 0.0003  \\
\hline
\multicolumn{5}{|c|}{300-W} \\
\hline 
VAE & 0.2349 &	0.2123 &	0.1450 &	0.0922 \\
SVI & 0.0012 &	0.0009	& 0.0006	& 0.0004 \\
PE-SVI-0 & 0.0798	& 0.0775	& 0.0697 &	0.0592  \\
PE-SVI-25 &  0.0011 &	0.0008 &	0.0007 &	0.0002  \\
\hline
\multicolumn{5}{|c|}{Synthetic} \\
\hline 
VAE & 78.6053 &	73.9616 &	66.1229 & 60.1511   \\
SVI & 0.0331 & 0.0117 & 0.0021 & 0.0005   \\
PE-SVI-0 & 0.9706 & 0.5561 & 0.5282 & 0.4197   \\
PE-SVI-25 &  0.0348 &	0.0119 & 0.0039 & 0.0027   \\
\hline
\multicolumn{5}{|c|}{SST} \\
\hline 
VAE & 0.4860&	0.4162&	0.3801&	0.3233 \\
SVI & 0.1411&	0.1228&	0.0895&	0.0506 \\
PE-SVI-0 & 0.3781&	0.3605&	0.3559&	0.3499 \\
PE-SVI-25 &  0.1417&	0.1229&	0.0887&	0.0517  \\
\hline
\end{tabular}
\end{center}

\caption{\label{table:a1}The results of experiments on Arch1. Refer to Section~\ref{sec:results} for discussion and analysis.}

\end{table}
\subsection{Methodology}

For all the datasets, we study the following feed-forward encoder-decoder or decoder-only architectures. For all the architectures, $|z|$ is the dimensionality of the latent space. The encoder has the same architecture as the decoder, only inverted. The following decoder architectures are used in this paper: [Arch1] $\mathcal{D}^{A1}_\theta(z): z \mapsto x$, [Arch2] $\mathcal{D}^{A2}_\theta(z): z \mapsto \mathbb{R}^{min(z \times 2,128)} \mapsto x$, [Arch3] $\mathcal{D}^{A3}_\theta(z): z \mapsto \mathbb{R}^{min(z \times 2,128)} \mapsto \mathbb{R}^{min(z \times 2,128)} \mapsto x$. All the models are ReLU activated. 
\begin{table}[t!]
\begin{center}
\small
%\fontsize{7}{10}\selectfont
\setlength{\tabcolsep}{6.2pt}
\renewcommand{\arraystretch}{1.4}
\begin{tabular}{l|c|c|c|c}
Model\ \textbackslash \ $|z|$ & 16 & 32 & 64 & 128 \\
\hline 
\multicolumn{5}{|c|}{CMU-MOSI} \\
\hline 
VAE & 0.5778&	0.3644 &	0.2767 &	0.2257 \\
SVI &  0.0642&	0.0170&	0.0020&	0.0015  \\
PE-SVI-0  & 0.0686&	0.0214	&0.0068&	0.0060  \\
PE-SVI-25 &  0.0644&0.0171	& 0.0020 & 	0.0019  \\
\hline
\multicolumn{5}{|c|}{300-W} \\
\hline 
VAE &  0.1711&	0.1279&	0.1090&	0.0489  \\
SVI & 0.0022	& 0.0014	& 0.0012 &	0.0010\\
PE-SVI-0 &  0.0698	& 0.0692 & 0.0669 & 0.0614  \\
PE-SVI-25 &   0.0047&	0.0042&	0.0031&	0.0020  \\
\hline
\multicolumn{5}{|c|}{Synthetic} \\
\hline 
VAE & 47.6520 & 29.7762 & 23.5845 & 17.8166  \\
SVI & 0.0940 & 0.0491 & 0.0172 & 0.0155  \\
PE-SVI-0  & 0.5445 & 0.5283 & 0.5242 & 0.4968  \\
PE-SVI-25 &  0.0730 &	0.0530 & 0.0292 & 0.0156   \\
\hline
\multicolumn{5}{|c|}{SST} \\
\hline 
VAE &  0.4552&	0.3994&	0.3040&	0.2576  \\
SVI & 0.1718	&0.1434&	0.1302&	0.0808 \\
PE-SVI-0 & 0.3951&	0.3624&	0.3268&	0.2417 \\
PE-SVI-25 & 0.1728&	0.1444&	0.1273&	0.0804  \\
\hline
\end{tabular}
\end{center}
%SOTA1=RMFN, SOTA2=MFN, SOTA3=MARN 
\caption{\label{table:a2}The results of experiments on Arch2. Refer to Section~\ref{sec:results} for discussion and analysis.}

\end{table}

The following models are studied in this paper: 

\textit{VAE:} Variational Auto-Encoder uses and encoder to perform posterior approximation and a decoder to reconstruct a given input. Encoder and decoder are trained together end to end. The amortization gap essentially may happen during training~\cite{cremer2018inference}. 

\textit{SVI:} We use Stochastic Variational Inference directly on free-form latent parameters. We make a mean-field assumption for amortizing the posterior approximation. The free parameters of the latent space are essentially the parameters of a Gaussian distribution. 

\textit{PE-SVI-0:} Essentially, this is the proposed model in this paper without the adjustment steps in Section \ref{sec:adjustment}. The latent inference is simply done using the trained encoder in Section \ref{sec:latenc}. 

\textit{PE-SVI-25:} This is essentially PE-SVI-0, with $25$ steps with adjusted learning rate as discussed in Section \ref{sec:adjustment}. 

For all the models, we assume no particular prior distribution for latent space, therefore, in this paper we are only concerned with expected likelihood under the approximate posterior distribution (first term of ELBo in Equation \ref{eq:elbo}). This essentially compares the models for their reconstruction power. Note, we do not argue that a good generative model has more properties than just good reconstruction; however, good reconstruction is required for good generative modeling. In theory, the second term in ELBo has no direct dependency on the reconstruction as it simply forces the latent space to follow a particular distribution. This term is the same for both SVI and VAE, and therefore, both models can be adapted to follow a particular desired latent space distribution. To compare the reconstruction performance of both models, we directly maximize the expected log-likelihood reconstruction term within ELBo, and report the negative of its value.

\subsection{Hyperparameter Space Search}\label{sec:hyper}

The VAE models in this paper are trained using Adam with learning rates $\{1,5,8\} \times 10e-\{2,3,4,5\}$ for a total of $3000$ epochs. SVI and PE-SVI models are trained using $10e-\{2,3\}$ for model parameters and $10e-\{1,2,3\}$ for latent parameters (model and latent learning rates are independent). The best models are picked based on the performance on validation-set, and directly applied to the test-set of each dataset (random $10\%$ held out for validation and test). The Reduced Adaptation Steps are a total of $25$ epochs and the learning rates of $\{1,5\} \times 10e-\{0,1,2\}$. The hyperparameter space is searched with $12 \times$ Tesla V100 gpus.

\section{Results and Discussion}\label{sec:results}
The results of experiments over all datasets, baselines and architectures are reported in Tables \ref{table:a1}, \ref{table:a2}, \ref{table:a3} respectively for Arch1,2,3. We report the observations from these tables as follows:

\textit{Performance Comparison (VAE, SVI):} Firstly, we report whether or not a gap exists between SVI and VAE performance. Tables \ref{table:a1}, \ref{table:a2}, \ref{table:a3} demonstrates superior performance for SVI over VAE, by a rather large margin in certain cases. This gap signals a performance suboptimality for VAE model, also observed in previous works~\citep{cremer2018inference,hjelm2016iterative}. 
\begin{table}[t!]
\begin{center}
\small
%\fontsize{7}{10}\selectfont
\setlength{\tabcolsep}{6.2pt}
\renewcommand{\arraystretch}{1.4}
\begin{tabular}{l|c|c|c|c}
Model\ \textbackslash \ $|z|$ & 16 & 32 & 64 & 128 \\
\hline 
\multicolumn{5}{|c|}{CMU-MOSI} \\
\hline 
VAE & 0.4900	&0.3623&	0.2180&	0.2771 \\
SVI &  0.0980&	0.0835&	0.0032&	0.0026 \\
PE-SVI-0 &  0.1062	& 0.0870	& 0.0061 &	0.0065 \\
PE-SVI-25 &  0.0987&	0.0837&	0.0033&	0.0027 \\
\hline
\multicolumn{5}{|c|}{300-W} \\
\hline 
VAE & 0.1466&	0.1142&	0.0742&	0.0351 \\
SVI & 0.0021	& 0.0013 &	0.0011 & 0.0009\\
PE-SVI-0 & 0.0489 & 0.0492& 0.0475&	0.0146 \\
PE-SVI-25 &  0.0046&	0.0041&	0.0030&	0.0020 \\
\hline
\multicolumn{5}{|c|}{Synthetic} \\
\hline 
VAE & 35.4315 &	29.5171 &	25.6991 & 25.4315\\
SVI & 0.1209 & 0.0937 & 0.0354 & 0.0185\\
PE-SVI-0 & 0.6167 & 0.5527 & 0.5262 & 0.4420 \\
PE-SVI-25 & 0.1181 & 0.0922 & 0.0310 & 0.0227 \\
\hline
\multicolumn{5}{|c|}{SST} \\
\hline 
VAE & 0.3618&	0.2966&	0.2008&	0.1611 \\
SVI & 0.2140&	0.1871&	0.1462&	0.1010 \\
PE-SVI-0 & 0.5243&	0.4633&	0.4871&	0.5589 \\
PE-SVI-25 &  0.3149	&0.1872&	0.1504&	0.1098  \\
\hline
\end{tabular}
\end{center}
%SOTA1=RMFN, SOTA2=MFN, SOTA3=MARN 
\caption{\label{table:a3}The results of experiments on Arch3. Refer to Section~\ref{sec:results} for discussion and analysis.}

\end{table}

\textit{Performance Comparison (VAE, PE-SVI-0):} Surprisingly, we observe that in majority of our experiments, PE-SVI-0 performs better than VAE. Both models use an identical encoder architecture to perform approximate posterior inference. However, the decoder training is different across the models. We suspect that the lack of a performance gap when training using SVI (in Early Decoder Training phase), allows subsequent training of the encoder (in Deferred Encoder Training phase) to be more successful; as compared to training with both which can essentially lead to suboptimal performance for both encoder and decoder. It should be noted that the ultimate purpose of PE-SVI is to reduce the steps required for SVI inference, and this comparison was made as byproduct of our experiments.

\textit{Performance Comparison (SVI, PE-SVI-25):}
PE-SVI-25, which performs 25 adjusted steps (see Section \ref{sec:adjustment}) after PE-SVI-0, is able to closely approximate the performance of the SVI model. For SVI, the number of required steps for inference convergence is usually higher than $1000$ across all our datasets. For example, convergence steps for SST Arch1 (Table \ref{table:a1}) with $|z|=128$ is $2381$ with learning rate of $0.001$ (non-convergent with $0.01$), while PE-SVI-25 reaches the same performance in $12$ steps (and plateaus afterwards) with learning rate of $0.1$. Thus higher learning rate (different than used for random initialization) can successfuly be used with PE-SVI, after Pace Adjustment. 

\textit{Performance Comparison (SVI, PE-SVI-0):} The comparison between SVI and PE-SVI-0 suggests the latent space learned by SVI is complex, and not perfectly reconstructable using an encoder, which naturally has limited inference capacity. Such a suboptimality naturally takes a toll at the training process~\citep{hjelm2016iterative}, as also observed from comparison between SVI and VAE. 

\textit{Performance Comparison (SVI across Arch1,2,3):} Surprisingly, depth of the decoder seems to negatively impact the performance of SVI. This demonstrates that in many cases, a small decoder may be enough to learn a generative model with good reconstruction. However, this depth positively impacts PE-SVI-0 and VAE, signaling that more powerful encoders are better capable at approximating the latent space learned by SVI. Regardless, PE-SVI-25 follows the performance of SVI. We did not observe any pattern in higher or lower number of epochs required for convergence of PE-SVI-25 across different depths. 

\textit{Performance Comparison ($|z|$):} Unanimously across all models, architectures and datasets, increasing the dimensionality of the latent space improves the performance. 

%\subsection{An Alternative Approach}
%Alternative to PE-SVI, one can train a VAE first, and subsequently refine it using SVI. This is similar to approach pursued by ~\cite{hjelm2016iterative} and  \cite{kim2018semi}\footnote{Note that \cite{kim2018semi} use Hessian methods to eliminate the need for SVI steps after VAE. However, the final objective of closing the gap is the same, regardless of using SVI or not.}. After VAE is trained, SVI is used in two ways: 1) without updating the decoder weights, just to update the parameters of the approximated posteriors, 2) updating both decoder weights and parameters of the approximate posteriors. Both of these are studied over CMU-MOSI dataset, and shown in Table BLAH. These subsequent SVI steps are adjusted using approach in Section \ref{sec:adjustment} and similar learning rates as Section \ref{sec:hyper}. However, instead of $25$ such updates, we allow for $100$ SVI steps. Results of Table BLAH suggest that in both cases (1) and (2) results slightly improve, but does not reach a similar performance as PE-SVI. We believe that this is due to the fact that approximate posterior parameters and decoder weights cannot easily recover from their initial suboptimal position (e.g. stuck in local minima) learned by VAE. 
\section{Conclusion}
In this paper, we presented a new approach called Pseudo-Encoded Stochastic Variational Inference (PE-SVI), to reduce the inference complexity of SVI during test time. Our approach relies on finding a suitable initial start point for gradient operations, which naturally reduces the required SVI steps. Furthermore, this suitable start point allows for taking larger SVI step sizes during test-time inference (compared to random initialization) which further reduces the required SVI steps. Essentially, we learn a parametric model to output this start point. In our experiments, PE-SVI achieves similar performance to SVI, however with a fraction of required inference steps. Furthermore, we observe that the initial PE-SVI start point (without any SVI steps) shows better performance than jointly training a decoder with an inference model (e.g. VAE).  

\nocite{langley00}

\bibliography{citations}
\bibliographystyle{icml2019}


%%%%%%%%%%%%%%%%%%%%%%%%%%%%%%%%%%%%%%%%%%%%%%%%%%%%%%%%%%%%%%%%%%%%%%%%%%%%%%%
%%%%%%%%%%%%%%%%%%%%%%%%%%%%%%%%%%%%%%%%%%%%%%%%%%%%%%%%%%%%%%%%%%%%%%%%%%%%%%%

\end{document}